\documentclass[
]{ceurart}

\usepackage{listings}
\usepackage{natbib}
\usepackage{pifont}
\usepackage{threeparttable}
\usepackage{tabularx}
\newcommand{\xmark}{\ding{55}}%
\begin{document}

\copyrightyear{2025}
\copyrightclause{Copyright for this paper by its authors.
  Use permitted under Creative Commons License Attribution 4.0
  International (CC BY 4.0).}

\conference{Workshop SIG Knowledge Management (FG WM) at KI2025, Potsdam, Germany}

\title{Ontology-Aligned Embeddings for Data-Driven Labour Market Analytics}

\author[1,2]{Heinke Hihn}[%
orcid=0000-0002-3244-3661,
email=heinke.hihn@iu.org,
]
\cormark[1]
\address[1]{IU International University of Applied Sciences, Department of Computer Science and Engineering, Berlin, Germany}
\address[2]{The Stepstone Group, AI Labs, Berlin, Germany}
\address[3]{The Stepstone Group, AI Labs, London, UK}
\address[4]{The Stepstone Group, AI Labs, Düsseldorf, Germany}

\author[2]{Dennis A. V. Dittrich}[%
orcid=0000-0002-4438-8276,
email=dennisalexivalin.dittrich@stepstone.com,
]

\author[3]{Carl Jeske}[%
email=carl.jeske@stepstone.com,
]

\author[3]{Cayo Costa~Sobral}[%
orcid=0000-0002-3899-4894,
email=cayo.costa-sobral@stepstone.com,
]

\author[4]{Helio Pais}[%
orcid=0000-0003-2352-298X,
email=helio.pais@stepstone.com,
]

\author[4]{Timm Lochmann}[%
email=timm.lochmann@stepstone.com,
]

\cortext[1]{Corresponding author.}

\begin{abstract}
The limited ability to reason across occupational data from different sources is a long-standing bottleneck for data-driven labour market analytics. Previous research has relied on hand-crafted ontologies that allow such reasoning but are computationally expensive and require careful maintenance by human experts. The rise of language processing machine learning models offers a scalable alternative by learning shared semantic spaces that bridge diverse occupational vocabularies without extensive human curation. We present an embedding-based alignment process that links
any free-form German job title to two established ontologies - the German \textit{Klassifikation der Berufe}
and the \textit{International Standard Classification of Education}. Using publicly available data from the German Federal
Employment Agency, we construct a dataset to fine-tune a Sentence-BERT model to learn the structure imposed by the
ontologies. The enriched pairs (job title, embedding) define a similarity graph structure that we can use for efficient approximate
nearest-neighbour search, allowing us to frame the classification process as a semantic search problem. This allows for greater flexibility, e.g., adding more classes. We discuss
design decisions, open challenges, and outline ongoing work on extending the graph with other ontologies and multilingual titles.
\end{abstract}


\maketitle

\section{Introduction}
\label{sec:introduction}
Occupational data is one of the most versatile bits of information about a subject in quantitative data and provides
a wide range of analytical use cases, such as socioeconomic indices, measures of workplace tasks, occupation-specific
health risks, gender segregation, and occupational closure~\cite{Christoph2020-yp}. Naturally, knowledge‐intensive
organisations in the employment sector aim to take advantage of this rich information, raising the need for systematic
and internationally comparable ontologies.

In the German labour market, job roles are codified by the \textit{Klassifikation der Berufe} (KldB 2010)~\cite{Bundesagentur-fur-Arbeit2021-ic,Dorpinghaus2023-xw},
while educational attainments follow \textit{Deutscher Qualifikationsrahmen für lebenslanges Lernen} (DQR)
~\cite{Unknown2013-wr,Unknown2024-ht}. The DQR is based on the \textit{European Qualification Framework}~\cite{mehaut2012european} and is highly
aligned with UNESCO's \textit{International Standard Classification of Education} (ISCED 2011)~\cite{UNESCO-Institute-for-Statistics2012-ot}.
Yet, these ontologies reside in separate silos and are rarely connected at scale, forcing practitioners to fall back on ad-hoc keyword heuristics that neither generalise nor
explain their recommendations.

Free-form job titles aggravate the problem as the same occupation may appear in different forms, e.g. ``Software Engineer'' or ``Software Developer''.
Recent studies have shown that dense vector representations help cluster these variants and
predict plausible career paths, but do not align the results to standard taxonomies suitable for
downstream reasoning~\cite{zhang2019job2vec,liu2022title2vec}.

This paper closes that gap by introducing a lightweight method that

\begin{itemize}
\item fine-tunes Sentence-BERT (SBERT)~\cite{reimers2019sentence} to predict 5-digit KldB 2010 codes with sub-second latency
\item infers corresponding ISCED 2011 ranges via rule-based heuristics, linking occupational and educational hierarchies
\item exposes the result as a lightweight graph for fast inference
\end{itemize}

The remainder of the paper reviews related work (Section~\ref{sec:related-work}), details data and methodology
(Section~\ref{sec:data-methodology}), presents empirical results (Section~\ref{sec:results}), and
concludes with a discussion and future research directions (Section~\ref{sec:discussion}).

\section{Related Work}
\label{sec:related-work}
\begin{table}
\caption{Hierarchical structure of occupational classification of the KldB 2010 codes.}
\centering
\begin{tabular}{@{}ll@{}}
\toprule
\textbf{Level} & \textbf{Example Denomination} \\
\midrule
10 Occupational Areas & 3: Occupations in construction, architecture, surveying and technical building services \\
37 Main groups (2-digit) & 32: Occupations in building construction above and below ground \\
144 Groups (3-digit) & 322: Civil engineering \\
702 Subgroups (4-digit) & 3225: Canal and tunnel construction \\
1300 Types (5-digit) & 32253: Canal and tunnel construction – complex specialist tasks \\
Specific job titles & 32253-100 Canal Construction Foreman (Kanalmeister/in) \\
\bottomrule
\end{tabular}
\label{tab:kldb_example}
\end{table}

Classification of job titles into various occupational and educational taxonomies has been a long-standing challenge in
labour market analytics. In the following, we summarise the state of the art, highlighting key contributions and
gaps that our work addresses.

\paragraph{Large Language Models (LLMs)} LLMs are increasingly combined with structured graphs to offer
both neural adaptability and symbolic transparency~\cite{VLDB2024LLMKG,Dehal2025KGLLM}. However,
they come with high computing costs, but the need for efficient methods is pressing: The Stepstone Group’s job
board hosts more than 600,000 open vacancies per year~\cite{stepstone2025numbers}, each with a free-form job title that
must be classified into
various occupational and educational taxonomies. This requires a lightweight, fast, and explainable solution that can
be integrated into real-time recommender systems~\cite{Dawson2021-za}, such as pay-band calibration~\cite{Clemens2014-uw},
labour force demand and supply analyses, job title normalisation~\cite{Zbib2022-zu}, or skill-gap detection~\cite{Brunello2021-tl}.

\paragraph{Automated occupation coding.} Early attempts to normalise free-text job titles for statistics relied on rule bases or
TF-IDF similarity. Transformer models now dominate the task.~\citeauthor{BaskaranMueller2023} fine-tuned a Sentence-BERT
variant on vacancy titles and achieved 0.86 accuracy at the 5-digit KldB 2010 level,~\citeauthor{Safikhani2023} compared BERT and GPT-3 on survey responses and showed that hierarchical loss functions boost KldB
coding by 15 pp over traditional baselines.

\paragraph{Embedding-enriched representations.} JobBERT introduced by~\citeauthor{Decorte2021} augments BERT with co-occurring ESCO skill labels to
learn dense job-title embeddings that outperform generic sentence encoders on ESCO normalisation. 

\paragraph{Skills and occupation knowledge graphs.} Beyond classification, several authors develop labour market knowledge
graphs.~\citeauthor{DeGroot2021} fuse job–skill relations from postings with ESCO/O*NET to drive skills-based matching and career-path
search. Seif et al. add temporal layers to capture how demand evolves, stressing graph update mechanisms.
Neither graph incorporates formal education codes or German-specific taxonomies.

\paragraph{Linking occupations to education.}~\citeauthor{Seif2024} proposed large-scale projections of qualification
supply and already use the fifth KldB 2010 digit (requirement level) as a proxy for educational attainment, but do not have an explicit ISCED 2011 mapping. Our approach bridges between KldB 2010 levels and ISCED
ranges to address this gap.

\paragraph{Positioning of this work.} The state-of-the-art covers (i) KldB 2010 or ISCO classification, (ii) skill-centric
knowledge graphs, and (iii) exploratory LLM-graph integrations, each in isolation. We propose a method
that combines occupational and educational codes, and exposes the result as
a queryable graph fit for downstream LLM applications, such as pay-band calibration or skill-gap detection.
By integrating occupational, educational, and managerial semantics in a single embedding space, our work extends prior
efforts and addresses an unmet need in knowledge-based workforce analytics.

\section{Data and Methodology}
\label{sec:data-methodology}
\begin{figure}
  \includegraphics[width=0.95\textwidth, trim={1cm 6cm 1.9cm 7.5cm}, clip]{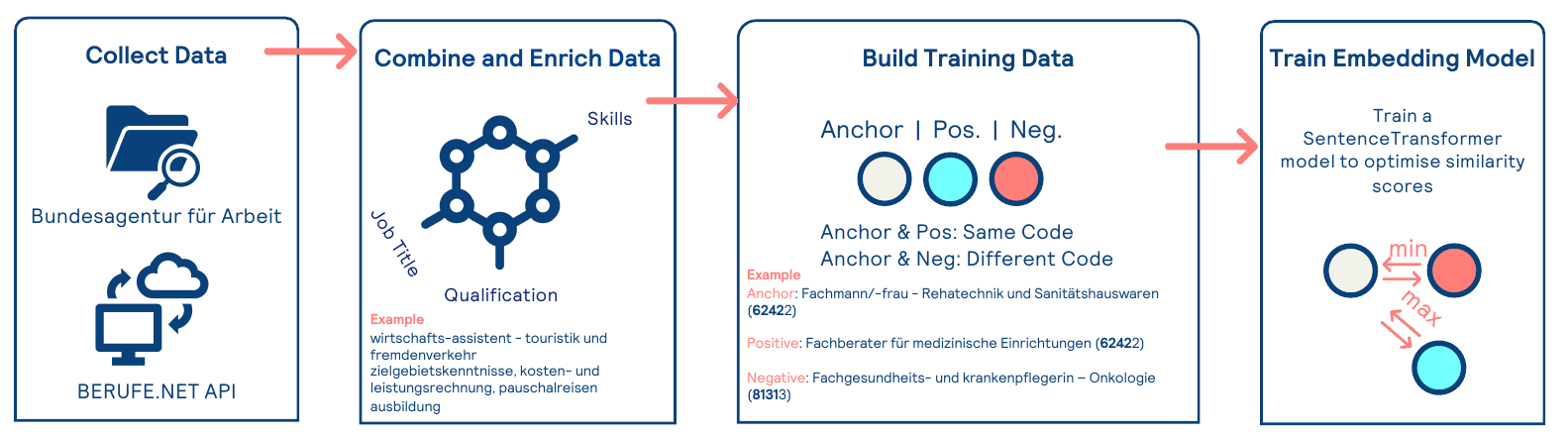}
  \caption{We collect publicly available data from the \textit{Bundesagentur für Arbeit}, enrich it, build a training data set,
  and fine-tune an embedding model. Anchor-Positive-Negative triplets are determined by matching KldB 2010 codes. In this example, we link to the anchor with KldB 62422 another job title from the KldB subgroup 6262 (``sales occupations (retail) selling medical supplies and healthcare goods'') as the positive and with a job title from KldB subgroup 8318 (``Occupations in nursing specialised in a particular branch of nursing'') as the negative sample.}
  \label{fig:exo_pipeline}
\end{figure}

This section details the resources, preprocessing steps, and modelling choices that enable the alignment of free-form German job titles with KldB 2010 and ISCED 2011.

\subsection{The German Classification of Occupations: KldB}
The \textit{Klassifikation der Berufe} (KldB) is a five-digit
code that uniquely identifies occupations based on tasks, skills, industry, and complexity developed and maintained
by the German Federal Employment Agency (\textit{Bundesagentur für Arbeit (BAfA)}). Each digit defines a level:
areas, main groups, groups, sub groups, and types. The last digit serves as an indicator for the complexity of the job:
1 identifies helper jobs that require no vocational training, 2 is for specialist activities, 3 are complex specialist activities, and 4 are highly complex activities.
Table~\ref{tab:kldb_example} gives an example. The fourth digit is an indicator for management duties: for requirement levels
2 and 3 it indicates supervisors, and for requirement level 4 it indicates managers. The distinction helps to
distinguish ``blue-collar'' and ``white-collar'' occupations.

We use an extended data set that contains not only official job titles, but also terms related to job titles. The data
contains all indexed key words supported by the BAfA job site. In total, it contains 525.207 terms.

\subsection{The International Standard Classification of Education: ISCED}
The International Standard Classification of Education (ISCED), maintained by the UNESCO Institute for Statistics \cite{UNESCO-Institute-for-Statistics2012-ot}, is the globally recognised taxonomy for describing programmes and attainment across all education systems - see Table \ref{tab:isced}. First introduced in 1976 and revised most recently in 2011, ISCED organises learning opportunities into a hierarchical code scheme (levels 0 to 8) that captures the progressive complexity and specialisation of study, from early childhood to doctoral research. Because each code is defined by internationally agreed content-based criteria rather than by national nomenclature, ISCED enables robust cross-country comparison, longitudinal tracking of educational expansion, and seamless integration of qualification data into labour market and demographic statistics.

\subsection{Training}
\label{subsec:training}
To train the embedding model, we select a baseline SentenceTransformer tuned on German from HuggingFace's sentence similarity
model repository~\cite{huggingfacegbert} by evaluating their performance on a fraction of the data. The training data set is constructed in the contrastive learning setting
of triplets~\cite{jaiswal2020survey}: Anchor - Positive - Negative. Positives and negatives are randomly assigned by KldB 2010 codes in relation to the
anchor sample, where positives share a KldB 2010 code, and negatives have different codes. We create two data sets: one by
the first four digits of the KldB 2019nd one by the last digit and then de-duplicate them. For each term in the base data
file, we randomly select three positives and negatives, resulting in roughly 1.5M training triplets.

The model is trained on a single GPU making use of several memory saving techniques~\cite{rojas2020study,soydaner2020comparison} to push the batch size as high as
possible, as this has been proven effective in contrastive learning and similar tasks~\cite{jaiswal2020survey,hihn2024online}.
We use Multiple Negatives Ranking Loss~\cite{henderson2017efficient} as a base loss and Matryoshka Loss~\cite{kusupati2022matryoshka} as
a meta-loss to encourage the model to learn embeddings on different dimensions (64 to 1024). Figure~\ref{fig:exo_pipeline}
gives an overview of the training process.

\begin{table}
  \caption{ISCED 2011 Levels~\cite{schneider2013international} define an encoding
  of educational attainments that makes different qualifications and requirements internationally comparable. Note, that codes 35\_2 and
  35\_3 were added to distinguish between forms of vocational training and are not part of the original ISCED 2011 levels. Note, that we only use a subset of the levels, excluding 01, 02, 10, 44, and 45.}
\centering
\begin{tabular}{@{}lll@{}}
\toprule
\textbf{Level} & \textbf{Educational Level} & \textbf{Example occupation} \\
\midrule
01, 02 & Pre-School & Not used in our approach \\
10 & Elementary School & Not used in our approach \\
24 & Lower secondary education & Demolition Worker \\
35\_2 & Two-year dual vocational training & Motor-Vehicle Service Assistant \\
35\_3 & Three-year dual vocational training & Automotive Mechatronics Technician – Vehicle Comm. Tech. \\
44, 45 & Post-secondary non-tertiary education & Not used in our approach\\
55 & Certified Vocational Specialist & Vocational Specialist – Foreign-Language Communication \\
64 & Bachelor's degree or eq. & Software Engineer \\
65 & Master Craftsperson or eq. & Master Craftsperson for Upholstery Technology \\
74 & Master's degree or eq. & Data Scientist \\
75 & Master Professional or eq. & Certified IT Technical Engineer \\
84 & Doctorate (Ph.D.) & University Professor \\
\bottomrule
\end{tabular}
\label{tab:isced}
\end{table}

\subsubsection{Query Structure}
To encourage the model to learn distinguishable parts of the embeddings, we define a query structure based on
special separator tokens. The query structure is defined as:
\begin{lstlisting}[language=Python]
query = "[JOB_TITLE_SEP] {job_title} [QUALIFICATION_SEP] {qualification} [SKILL_SEP] {skills}",
\end{lstlisting}
where the separator tokens are special tokens defined in the tokenizer and are thus not part of the vocabulary on which
the model is trained.
\newpage
Skills and qualification are extracted through the publicly available BERUFENET.API~\cite{fischer2025berufenet} which provides a RESTful interface to
the BAfA's occupational database. An example of a set of skills is shown here (translated into English):
\vspace*{1em}
\newline
\begin{tabular}[t]{@{}l p{0.75\linewidth}@{}}
  \textbf{Job Title}: Finance Assistant &
  \textbar{} \textbf{KldB}: 76911 \textbar{} \textbf{Skills}:\\[2pt]
  \multicolumn{2}{@{}p{\linewidth}@{}}{\small%
    international business; construction/mortgage financing; home-savings
    business; credit assessment; financial-services consulting; financial
    planning; costing; lending operations; customer consulting/support;
    marketing; retail banking; savings and investment services; tax law;
    insurance business; insurance law; contract processing/administration;
    securities business; field service/external sales; foreign-trade
    financing; banking and stock-exchange law; banking and capital-markets
    law; controlling; electronic banking; corporate banking; account
    management; cost and performance accounting; claims processing;
    sustainable investments; online banking; auditing/internal audit;
    clerical processing; contract law; payment transactions; competence group
    bank products; competence group office communication (MS-Office);
    competence group insurance lines}%
\end{tabular}
\vspace*{1em}

To simplify training, we use only a subset of the available qualifications, namely: Helper Jobs, Vocational Training,
Additional Vocational Training, University Degree, Civil Servants, and Armed Forces Personnel. The remaining qualifications
are assigned to one of them, e.g., ``Study Programme'' is assigned to University Degree - see Table \ref{tab:qualifications} for the full mapping we use. In cases where the job title
is associated with management duties, as indicated by the fourth digit being $9$, we add ``with management duties'' to the qualification, for example, ``university degree
with management duties''. 

\subsubsection{Enriching with ISCED}
\begin{table}
  \centering
  \caption{Mapping of original occupational qualification values to consolidated groups for simplified grouping during training.}
  \begin{tabularx}{\textwidth}{@{}l X@{}}
    \toprule
    \textbf{Consolidated group} & \textbf{Original categories mapped to it} \\
    \midrule
    Weiterbildung & Andere Weiterbildungen; Kaufmännische Weiterbildungen; Tätigkeiten nach Weiterbildung; Weiterbildungen (bedingen Hochschulstudium) \\
    Ausübungsformen & Ausübungsformen \\
    Bamtenausbildung & Beamtenausbildungen; Tätigkeiten nach Beamtenausbildung \\
    helfer-/anlerntätigkeiten & Helfer-/Anlerntätigkeiten \\
    Meister & Meister \\
    soldatenausbildung & Soldatenausbildungen; Tätigkeiten nach Soldatenausbildung \\
    Ausbildung & Sonstige Ausbildungen; Tätigkeiten nach Ausbildung; \\
    & Duale Ausbildung,  Ausbildungen für Menschen mit Behinderungen; Berufsfachschulausbildungen (rechtlich geregelt)  \\
    Studium & Studienfächer/-gänge; Tätigkeiten nach Studium \\
    Techniker & Techniker \\
    \bottomrule
  \end{tabularx}
  \label{tab:qualifications}
\end{table}

The BAfA provides high-level qualification levels,
but does not map them to ISCED codes. Regulated occupations and academic
degrees are classified within the German Qualifications Framework for
Lifelong Learning \cite{Unknown2013-wr, Unknown2024-ht} into a set of DRQ
levels that can be directly mapped to ISCED levels. We therefore define
a set of rules to map the KldB codes to ISCED levels. The mapping is
based on a combination of the requirement level, the qualification
provided by the BAA, the job title, and, if it exists, its corresponding
DQR level and in some cases the KldB code itself. For example, if the
job title contains an education level such as ``Bachelor of Science'' or
``staatlich geprüfter Betriebswirt'', we map it to the corresponding ISCED
levels 64 and 65 - see Table~\ref{tab:isced_mappings} for more examples of
the mapping.

\subsection{Inference}
\label{sec:inference}
\begin{table}
\caption{Examples of how qualification, requirement level, and keywords are mapped to ISCED 2011 levels.}
\centering
\begin{tabular}{@{}llll@{}}
\toprule
\textbf{ISCED} & \textbf{Qualification} & \textbf{Req. Level} & \textbf{Example Keywords}\\
\midrule
24 & Helper jobs & 1 & Trainee (Auszubildende/r) \\
35\_2 & Vocational Training & 1, 2 & official job titles  \\
35\_3 & Vocational Training & 2, 3 & official job titles \\
55 & Additional Vocational Training & 3 & Vocational Specialist (Berufsspezialist/in) \\
64 & University Degree & 3 & \xmark\\
65 & Additional Vocational Training & 3 & Master Craftsperson (Meister/in) \\
74 & University Degree & 4 & \xmark \\
75 & Additional Vocational Training & 4 & Certified IT Technical Engineer \\
84 & University Degree & 4 & Professor\\
\bottomrule
\end{tabular}
\label{tab:isced_mappings}
\end{table}

Inference is performed by a $k$-Nearest-Neighbour ($k$-NN) search on the embedding space. The $k$-NN search is performed
using the HNSW library~\cite{malkov2018efficient,malkov2023hnswlib} which allows for an efficient approximate nearest-neighbour search
in logarithmic time complexity. The final decision on the $k$-NN search is made by a majority vote of the top-$k$-nearest
neighbours, where $k$ is a hyperparameter that can be tuned according to the desired level of granularity. The majority vote
determines the KldB 2010 code and the ISCED 2011 level separately.

\subsubsection{Inference Examples}
For the query Construction Supervisor (Baustellenleitung) we get the following results:

\begin{center}
\begin{tabular}{@{}llll@{}}
\textbf{$k$-nearest} & \textbf{ISCED} & \textbf{Type} & \textbf{CS}\\
Construction Supervisor (Bauaufseher)          & 74, 75 & 31194 & 0.945 \\
Construction Supersivor (Bauaufseherin)        & 74, 75 & 31194 & 0.942 \\
Contruction Manager (Baustellen-Manager)   & 74, 75 & 31194 & 0.939,
\end{tabular}
\end{center}
where CS is the cosine similarity between the query and the item. Combing them with a majority vote results in ISCED 2011 74, 75
(Master Degree, Master Professional) and KldB 31194
(Managers in construction scheduling and supervision, and architecture). The requirement level 4 of the nearest job titles
result in high educational levels. The fourth digit being 9 indicates jobs with managerial duties.

Adding skills ``Bathroom planning, Customer Management,
Sanitary Engineering, Personnel Responsibility'' (``Badplanung, Kundendienst, sanitärtechnik, personalverantwortung'')
specifies the query further and results in:

\begin{center}
\begin{tabular}{@{}llll@{}}
  \textbf{$k$-nearest} & \textbf{ISCED} & \textbf{Type} & \textbf{CS}\\
Sanitary Engineering Manager (Sanitärtechnikmanager)          & 65, 75 & 34293 & 0.931 \\
Sanitary Engineering Manager (Sanitärtechnikmanagerin)        & 65, 75 & 34293 & 0.928 \\
Installation Supervisor (Assembly) (Installationsleiterin (montage))  & 35\_3, 55, 65, 75 & 31193 & 0.921
\end{tabular}
\end{center}

Again, combining them with a majority vote result in ISCED 2011 65, 75 (Bachelor Professional, Master Professional), and
KldB 34293 (supervisors in sanitation, heating, ventilation, and air conditioning). Adding specific skills uncovers
jobs relevant to the task of sanitary engineering and management. The requirement level 3 indicates a specialist job, which
requires additional specialist vocational training. This reclassification also shifts the title from ``Manager'' to
``Supervisor'', underscoring its hands-on nature as a functional specialist lead role that relies on advanced vocational training
rather than purely managerial duties.

\section{Results}
\label{sec:results}
\subsection{Model Evaluation}
\label{sec:model_evaluation}
A comparative evaluation is particularly difficult as there is no common evaluation data set or procedures for KldB
classification. To facilitate a quantitative and standardised evaluation we compare our method to a $k$-NN classifier fit on
a variety of pre-trained models on occupational data. We acknowledge that the comparison may not be fair, as only our
model has been explicitly trained on KldB data, but as the results suggest training on ESCO and even a simple German
corpus provides a decent baseline. Furthermore, we compare against an TF-IDF embedding~\cite{sparckjones1972statistical}
both as a basis for a $k$-NN classifier and as features for a logistic regression classifier. We also evaluate the retrieval
accuracy of the embedding models by inspecting their respective \textit{mAP@3} and \textit{mAP@5} metrics.

In the requirement-level classification, we take a closer look at the predictions on a class level and compare our
results with one other method. Note, that the model we use for comparison was trained to classify full job ads \cite{krueger_inpress_explainable_ai}, so it's reported performance here may not reflect its capabilities on the full information provided by the complete job ad.

\begin{figure}
  \includegraphics[width=\textwidth]{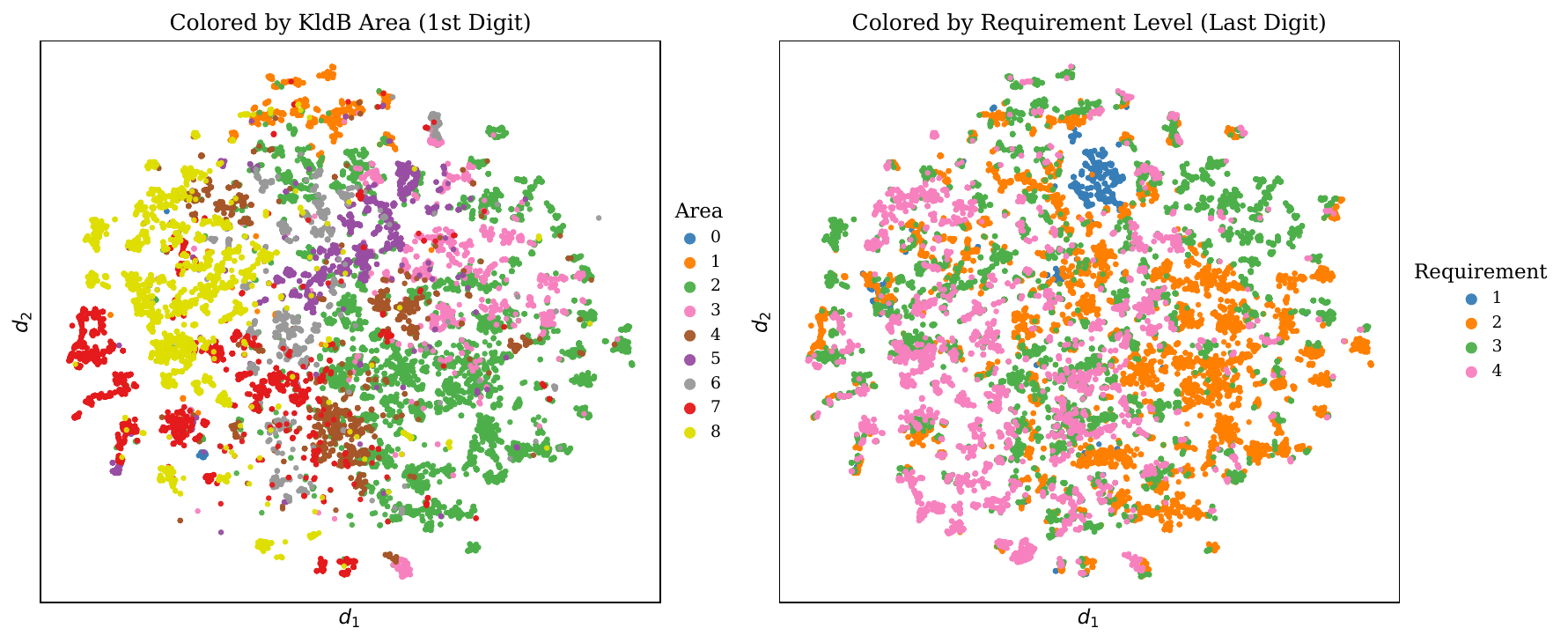}
  \caption{2D -projection using t-SNE~\cite{vanDerMaaten2008} of the embedding of the test dataset learned by our model,
  showing the structure it has learned for KldB 2010 areas (left) and requirement levels (right). We can see clusters evolving - especially in the requirement level groupings.}
  \label{fig:embedding_vis}
\end{figure}

The data set is divided into 80\% for training and 20\% for validation and testing. The metrics are computed per class and
aggregated by the macro-average, giving equal weight to the minority and majority classes. For testing when using evaluation
data from official sources, we have found that $k=1$ is optimal. The results of the evaluation on the test/train split
with $k=1$ are shown in Table~\ref{tab:knn_eval} and we show a low-dimensional visualization of the embedding space in
Figure~\ref{fig:embedding_vis}.

\begin{table}
  \centering
  \caption{Evaluation metrics (macro-F$_1$, rounded to three decimals). Baselines use TF–IDF or pretrained embeddings
  on occupational ontologies~\cite{rosenberger2025careerbert,jobbert2021decorte}. The best result in a setting is
  indicated by a \textbf{bold} typeface.}
  \setlength\tabcolsep{6pt}
  \begin{tabular}{lcccccc}
    \toprule
    \multicolumn{7}{c}{\textbf{Classification Evaluation}}\\
      \midrule
    \multirow{2}{*}{\textbf{Level}} &
    \textbf{LogReg + } &
    \multicolumn{4}{c}{\textbf{$k$-NN +}} &
    \textbf{Our Approach} \\
    \cmidrule(lr){2-2} \cmidrule(lr){3-6} \cmidrule(lr){7-7}
    & TF–IDF$^{\ddag\ddag}$ & TF–IDF$^{\ddag\ddag}$ & \texttt{gBERT}$^{\dag}$ & \texttt{CareerBERT}$^{\ddag}$ & \texttt{JobBERT}$^{\ddag}$ & $64$ dim.  \\
    \midrule
    Area        & $0.910$ & $0.960$ & $0.955$ & $0.950$ & $0.940$ & $\mathbf{0.976}$ \\
    Main Group  & $0.898$ & $0.942$ & $0.943$ & $0.944$ & $0.917$ & $\mathbf{0.968}$ \\
    Group       & $0.850$ & $0.923$ & $0.906$ & $0.914$ & $0.884$ & $\mathbf{0.948}$ \\
    Sub-Group   & $0.717$ & $0.888$ & $0.848$ & $0.866$ & $0.846$ & $\mathbf{0.926}$ \\
    Type        & $0.632$ & $0.858$ & $0.778$ & $0.792$ & $0.780$ & $\mathbf{0.886}$ \\
    Requirement & $0.852$ & $0.923$ & $0.900$ & $0.903$ & $0.928$ & $\mathbf{0.961}$ \\
    \bottomrule
    \toprule
    \multicolumn{7}{c}{\textbf{Retrieval Evaluation}} \\
    \midrule
    \multirow{2}{*}{\textbf{Level}} &
    \multicolumn{3}{c}{\textbf{mAP@3}} &
    \multicolumn{3}{c}{\textbf{mAP@5}} \\
    \cmidrule(lr){2-4} \cmidrule(lr){5-7}
    & \texttt{CareerBERT}$^{\ddag}$ & \texttt{JobBERT}$^{\ddag}$ & Ours ($64$ dim.) & \texttt{CareerBERT}$^{\ddag}$ & \texttt{JobBERT}$^{\ddag}$ & Ours ($64$ dim.)  \\
    \midrule
    Area        & $0.967$ & $0.951$ & $\mathbf{0.981}$ & $0.960$ & $0.940$ & $\mathbf{0.979}$ \\
    Main Group  & $0.953$ & $0.929$ & $\mathbf{0.971}$ & $0.944$ & $0.918$ & $\mathbf{0.968}$ \\
    Group       & $0.938$ & $0.884$ & $\mathbf{0.960}$ & $0.928$ & $0.899$ & $\mathbf{0.956}$ \\
    Sub-Group   & $0.917$ & $0.888$ & $\mathbf{0.946}$ & $0.905$ & $0.876$ & $\mathbf{0.939}$ \\
    Type        & $0.888$ & $0.861$ & $\mathbf{0.929}$ & $0.874$ & $0.847$ & $\mathbf{0.919}$ \\
    Requirement & $0.947$ & $0.946$ & $\mathbf{0.970}$ & $0.936$ & $0.937$ & $\mathbf{0.964}$ \\
    \bottomrule
  \end{tabular}
  \begin{itemize}[flushleft]
    \footnotesize
    \item[\dag] 1024-dimensional sentence embeddings trained on a German retrieval corpus.
    \item[\ddag] 768-dimensional sentence embeddings trained on the ESCO ontology\cite{robert2014esco}.
    \item[\ddag\ddag] Uses $(5,3)$-grams.
  \end{itemize}
  \label{tab:knn_eval}
\end{table}

\begin{table}
  \centering
  \caption{Evaluation metrics for the requirement level (5th digit) of the KldB 2010 codes. We compare our results
  with the classification model \texttt{oja\_reqlevel\_de} (OJRD)~\cite{huggingfaceojareqlevel,krueger_inpress_explainable_ai} and
  the averaged results.}
  \setlength\tabcolsep{6pt}
  \begin{tabular}{lcccccc}
    \toprule
    \multirow{2}{*}{\textbf{Requirement Level}} &
    \multicolumn{2}{c}{\textbf{Precision}} &
    \multicolumn{2}{c}{\textbf{Recall}} &
    \multicolumn{2}{c}{\textbf{F1}} \\
    \cmidrule(lr){2-3} \cmidrule(lr){4-5} \cmidrule(lr){6-7}
    & Ours & OJRD & Ours & OJRD & Ours & OJRD \\
    \midrule
    1: Helper Jobs                    & $\mathbf{0.963}$ & 0.319 & $\mathbf{0.963}$ & 0.536 & $\mathbf{0.963}$ & 0.400 \\
    2: Specialist Activities          & $\mathbf{0.983}$ & 0.551 & $\mathbf{0.979}$ & 0.799 & $\mathbf{0.981}$ & 0.652 \\
    3: Complex Specialist Activities  & $\mathbf{0.951}$ & 0.605 & $\mathbf{0.960}$ & 0.441 & $\mathbf{0.956}$ & 0.512 \\
    4: Highly Complex Activities      & $\mathbf{0.959}$ & 0.675 & $\mathbf{0.942}$ & 0.448 & $\mathbf{0.946}$ & 0.533 \\
    \midrule
    & Ours & OJRD & & & & \\
    Macro Averaged F1 & $\mathbf{0.961}$ & 0.525 & & & &\\
    \bottomrule
  \end{tabular}
  \label{tab:req_level_eval}
\end{table}
\subsection{Ablation Experiments}
\label{subsec:ablation}

To gauge how much the classifier relies on surface forms rather than true
semantic content, we conducted four targeted ablation studies.  Each study
starts from a \emph{Original} test slice and rewrites the titles according to
a specific perturbation rule. All rewritten titles remain plausible German phrases, so
no out‐of‐domain noise is injected. Table~\ref{tab:ablation_summary} summarises the macro metrics.

\paragraph{Management nouns {\small(Leiter\,/\,Leiterin\,/\,Leiter/in $\rightarrow$ Leitung).}}
Replacement of the gendered or slash‐notation head noun with its gerundive form
(\textit{Leitung}) affects 765 titles.  The subgroup–level accuracy drops by
1.7 pp (0.968 $\rightarrow$ 0.951), while the requirement digit loses only
0.4 pp.  Thus, the embedding is largely insensitive to the noun morphology
often used to encode gender.

\paragraph{Gender Inflection}
On 3,324 neutral titles, we generated masculine and feminine variants.
Performance declines by less than 0.6 pp at the subgroup level and remains
virtually unchanged for the requirement digit, confirming that the model
handles gender‐specific suffixes (\textit{–er} vs.\ \textit{–in}) robustly.

\paragraph{Word‐Order Reversal}
For 1,005 two‐token titles we swapped the token order
(\textit{chemietechnischer Assistent} $\mapsto$
\textit{Assistent chemietechnischer}).  Because German occupational
compounds are mainly head‐final, this permutation removes a strong lexical
cue.  Accuracy drops by 1.9 pp at the subgroup level and 1.4 pp at requirement,
but stays above 0.85, indicating that contextual embedding still recovers
much of the semantics.

\begin{table}
  \centering
  \caption{Macro‐averaged performance (Sub-Group = 4-digit, Requirement = 5th digit)
  under three perturbation scenarios.  \,$\Delta$ denotes the absolute change
  relative to the original slice.}
  \label{tab:ablation_summary}
  \begin{tabular}{@{}lcccccc@{}}
    \toprule
    \multirow{2}{*}{\textbf{Perturbation}} & \multicolumn{3}{c}{\textbf{Sub-Group (4-digit)}} &
    \multicolumn{3}{c}{\textbf{Requirement (5th digit)}}\\
    \cmidrule(lr){2-4}\cmidrule(l){5-7}
     & Acc.\ & F1 & $\Delta$Acc.\,[pp] & Acc.\ & F1 & $\Delta$Acc.\,[pp]\\
    \midrule
    Management noun & 0.951 & 0.952 & -1.7 & 0.982 & 0.981 & -0.4 \\
    Gender variant  & 0.866 & 0.863 & -0.6 & 0.950 & 0.949 & -0.3 \\
    Word order      & 0.858 & 0.862 & -1.9 & 0.881 & 0.882 & -1.4 \\
    \bottomrule
  \end{tabular}
\end{table}

Across the three scenarios, the classifier retains at least 0.85 accuracy at the
subgroup level and more than 0.98 at higher aggregation levels.  The
greatest vulnerability arises when the syntactic head is moved to sentence
initial position, reinforcing the intuition that the model encodes head
nouns more strongly than modifiers.  Still, the modest degradation confirms
that the embedding space has learned a degree of permutation invariance that
is indispensable for noisy inputs in the real world.

\section{Discussion and Future Work}
\label{sec:discussion}
We have presented a lightweight, and embedding-based knowledge representation system that aligns free-form German job
titles with the KldB 2010 and ISCED 2011 ontologies, enabling data-driven labour market analytics. Designed for
real-time recommenders, the system is fast, memory efficient and fully explainable.

Across the KldB 2010 hierarchy the system attains high macro-accuracy and F1 scores, with peak performance at the area
and main-group levels. Performance gradually degrades toward the type level, which is expected given the greater
lexical variability of job titles and the smaller number of cases available at the more granular classification levels.
Despite this, the model performs strongly on the requirement dimension, showing that it still captures much of the
semantic complexity of job roles.

\subsection{Future Work}
\label{sec:future_work}
Currently, the system uses the KldB 2010 codes to infer ISCED 2011 levels, but this could be inverted to
improve the KldB predictions.  This could be achieved by
introducing a two-step prediction process: first, predicting the ISCED 2011 level based on the job title and
then using the predicted ISCED level to refine the KldB 2010 prediction.
This would allow the system to take advantage of the educational hierarchy to improve the accuracy of KldB predictions, especially
for job titles that are ambiguous or have multiple interpretations.

The system is easily extensible: new ontologies, languages, or task-specific modules can be plugged in with minimal
engineering effort. The planned extensions include (i) adding ESCO/ISCO codes and multilingual job titles to widen the global
applicability, (ii) replacing the current rule-based ISCED 2011 mapper with a learned component, and (iii) revisiting
the training algorithm to make the KldB 2010 hierarchy explicit, which should bolster performance at the type level.

\subsection{Evaluation Challenges and Opportunities}
Reliable evaluation remains difficult because there is no publicly available, gold-standard set of free-form titles
mapped to all 1300 KldB 2010 codes. Although we can benchmark on the subset of titles already present in official datasets,
truly open-ended job titles still require expert annotation, which is costly and prone to subjective disagreement.
A promising alternative is controlled access to social security contribution data, where employers have already
supplied both job titles and KldB 2010 codes. Such administrative data would provide a much larger quasi-gold corpus
against which the accuracy and fairness of the system could be measured.

\begin{acknowledgments}
The authors thank the German Federal Employment Agency (BAfA) for providing the data and BERUFENET.API, which enables us to
access the occupational database. The authors also thank the Stepstone Group for supporting this research and providing the
necessary resources. The authors acknowledge the contributions of our colleagues at Stepstone Group AI Labs, who provided valuable feedback and
insights during the development of this work. We also thank the contributors to the HuggingFace model repository for providing the
baseline Sentence-BERT model that we used for fine-tuning. Lastly, the authors thank Kai Krüger from the Federal Institute of Vocational Training and Education for providing additional information about the \texttt{oja\_reqlevel\_de} model used for evaluation.
\end{acknowledgments}

\section*{Declaration on Generative AI}
During the preparation of this work, the authors used LLMs for grammar and spelling checks, summarising, and rephrasing.
After using these tools/services, the authors reviewed and edited the content as needed and take full
responsibility for the publication's content.

\bibliography{main.bib}


\end{document}